\documentclass[runningheads]{llncs}
\usepackage{graphicx}
\usepackage{amsmath}
\usepackage{amsfonts}
\usepackage{subfig}
\usepackage{multirow}
\usepackage[ruled,linesnumbered]{algorithm2e}
\usepackage{times}

\newcommand{\ME}{\mathcal{E}}
\newcommand{\ML}{\mathcal{L}}

\newcommand{\cluster}{\textsc{Cluster}}
\newcommand{\name}{\textsc{Name}}
\newcommand{\encode}{\textsc{Encode}}
\newcommand{\session}{\textcircled{s}}
\newcommand{\avg}{\textsf{avg}}

\newcommand{\onlyPaper}[1]{}
\newcommand{\onlyReport}[1]{#1}

\begin{document}
\title{From Low-Level Events to Activities: \\ A Session-Based Approach \onlyReport{\\(Extended Version)}}

\author{Massimiliano de Leoni\inst{1} \and Safa Dundar\inst{2}}

\institute{Department of Mathematics\\ University of Padua\\Via Trieste, 63 - 35123 Padua (Italy)
\and Eindhoven University of Technology, Eindhoven, The Netherlands \\
\email{deleoni@math.unipd.it}}

\maketitle              
\begin{abstract}
Process-Mining techniques aim to use event data about past executions to gain insight into how processes are executed.
While these techniques are proven to be very valuable, they are less successful to reach their goal if the process is flexible and, hence, it exhibits an extremely large number of variants. Furthermore, information systems can record events at very low level, which do not match the high-level concepts known at business level.
Without abstracting sequences of events to high-level concepts, the results of applying process mining (e.g., discovered models) easily become very complex and difficult to interpret, which ultimately means that they are of little use.
A large body of research exists on event abstraction but typically a large amount of domain knowledge is required, which is often not readily available. Other abstraction techniques are unsupervised, which ultimately return less accurate results and/or rely on stronger assumptions.
This paper puts forward a technique that requires limited domain knowledge that can be easily provided.
Traces are divided in sessions, and each session is abstracted as one single high-level activity execution. The abstraction is based on a combination of automatic clustering and visualization methods. The technique was assessed on two case studies about processes characterized by high variability. The results clearly illustrate the benefits of the abstraction to convey accurate knowledge to stakeholders.
\end{abstract}
\section{Introduction}
Nowadays, large, complex organizations leverage on well-defined processes to try to carry on their business more effectively and efficiently than their competitors.
In a highly competitive world, organizations aim to continuously improve their business performance, which ultimately boils down to improving their process.

The first step towards improvement is to understand how processes are actually being executed. The understanding of the actual process enactment is the goal of process mining. This research field focuses on providing insights by reasoning on the actual process executions, which are recorded in so-called event logs~\cite{Aalst.2016}. Event logs group process events in traces, each of which contains the events related to a specific process-instance execution. An event refers to the execution of an activity (e.g., \emph{Apply for a loan}) for a specific process instance (e.g.\ customer \emph{Mr.\ Bean}) at a specific moment in time (e.g.\ on January, 1st, 2018 at 3.30pm).

\begin{figure}[t!]
  \centering
  \includegraphics[width=1\textwidth]{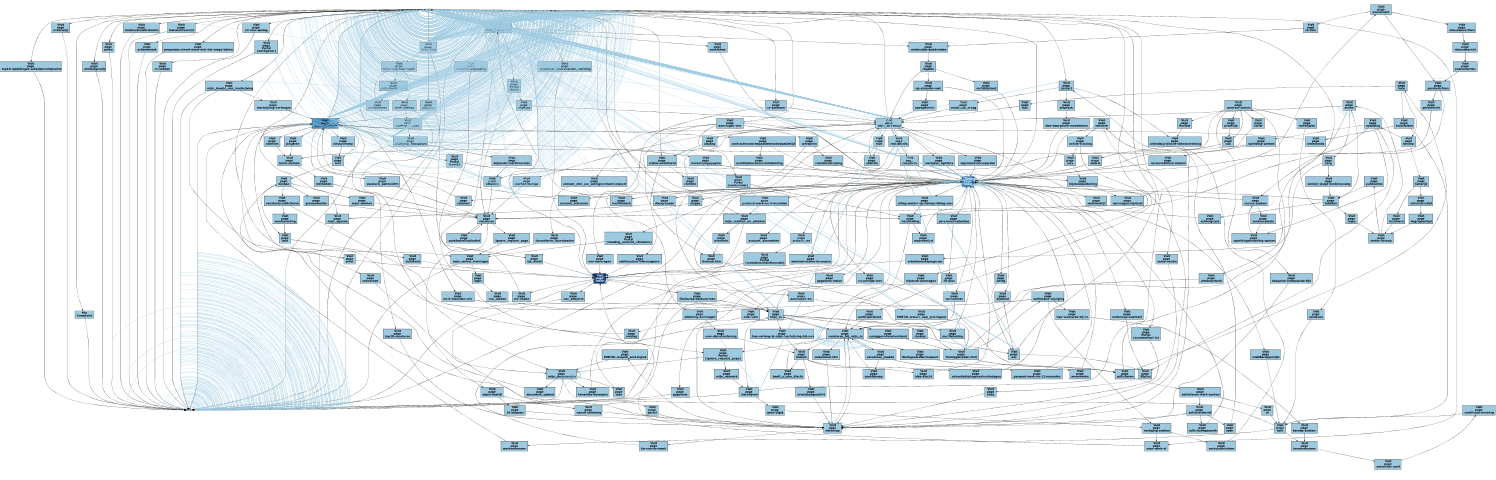}
  \caption{A model for a very flexible process, which shows an ocean of variability.}
  \label{fig:UWVNoAbstractionModel}
\end{figure}

While process mining has proven to be effective in a wide range of application fields, it has shown its limitation when the process intrinsically allows for a high degree of flexibility~\cite{Aalst.2016}, or information systems record executions into logs where events are at a lower-level granularity than the concepts that are relevant from a business viewpoint. Both of the problems lead to an ``ocean'' of observed process behavior.
This means that, e.g., if one tries to discover a process model, one obtains a model that is very complex and/or low-level, thus being difficult to interpret. As a matter of fact, if the granularity is too low level, even the event-log visualization through dotted charts~\cite{Aalst.2016} is less useful: users are confronted with a chart with too many dots to draw insightful conclusions.
\onlyPaper{Figure~\ref{fig:UWVNoAbstractionModel} shows an example of a model that was discovered from an event log that records huge behavioral variabilities. The model was obtained through the new Heuristic Miner~\cite{Mannhardt.2017e} and refers to the page-visit behaviour of the \emph{werk.nl} web site.}

Extreme complexity and difficulty of interpretation contrast the initial purpose of process mining: conveying interpretable insights and knowledge to process stakeholders and owners.
Typical examples are in health-care\onlyReport{~\cite{mannhardt2016low}}, customer journey\onlyReport{~\cite{Terragni:2019:OCJ:3297280.3297288}}, on-line retailer shops, supermarkets, hospitals, home automation, and IoT systems.

Similarly to existing related work (see Section~\ref{sec:relatedwork}), here we advocate the need of abstracting low-level events to high-level activities. However, differently from existing related work, we do not want to rely on the provision of an extensive amount of domain knowledge as existing approaches require: this can be hard in several domains. On the other hand, we want to avoid completely unsupervised approaches, which naturally show lower accuracy and/or rely on strong assumptions.

To balance accuracy and practical feasibility, we aim at a technique that requires process analysts to only feed in knowledge that is limited in quantity and easy to provide.
In a nutshell, the idea is that events of the same trace can be clustered into sessions such that the time distance between the last event of a session and the first event of the subsequent session is larger than a user-defined threshold.
Each trace is seen as a sequence of sessions of events.
These sessions are encoded into data points to be clustered; this way, each session is assigned to one cluster. The abstract event log is created such that the entire session is replaced by a high-level event that indicates to which cluster the session belongs. The high-level events needs to be named: The centroids of the clusters provide meaningful information for a process stakeholder to identify the high-level activity that corresponds to each cluster. To support stakeholder in this identification, visualization techniques are foreseen, based on heat maps. However, the latter is optional: e.g., without domain knowledge, each cluster may be given a name that coincides with that of the most frequent activity in the sessions of the cluster, or with a concatenation of the names of those most frequent, if more than one clearly stand out.

The benefit and feasibility of the proposed technique was assessed on two real-life case studies. The first refers to the \url{www.werk.nl} web site. Results show that overcomplex, low-level process models can be converted into high-level counterparts that are accurate according to the process-mining metrics, and that are simply enough to be able to convey information that has business value.\footnote{Here, a process is intended as a set activities that are executed while complying given ordering constraints. The activities can be of any nature, ranging from those more traditional performed, e.g., by a bank or city-hall employee, till web-page visits or those executed by domotics or IoT systems, such as by/with TVs, ovens, bulbs, bath tubes, or heaters.}
However, the idea of a session-based clustering goes beyond analysing web sites; it certainly applies to other domains, including on-line retailer shops, supermarkets, hospitals, home automation, and IoT systems. In general, one can apply the proposed technique to any domain in which events happen in batches/sessions. A second case study showcases the wider applicability of the technique and focuses on the executions of a process to manage building-permit requests.

\emph{The technique is not only beneficial when discovering a model, but also in a wider range of applications of diverse process-mining techniques.} As a support of this statement, we showcase an example: for the second case study, the abstract log is used to compare the management of building permits when different city-hall employees are responsible.

\onlyReport{Section~\ref{sec:example} introduces the initial motivating example of the \url{www.werk.nl} web site.}
Section~\ref{sec:clustering} introduces the abstraction technique, while Section~\ref{sec:eval} reports on the evaluation on the two cases. Section~\ref{sec:relatedwork} compares with the related work while Section~\ref{sec:conclusion} concludes the paper, delineating the avenues of future work.

\onlyReport{\section{Motivating example}
\label{sec:example}

The \url{www.werk.nl} web site is a very significant example of customer journey, intended as the product of the interaction between an organization and a customer throughout the duration of their relationship. Gartner highlights the importance of managing the customer's experience, which is seen as ``the new marketing battlefront''.\footnote{Key Findings From the Gartner Customer Experience Survey - \url{https://www.gartner.com/smarterwithgartner/key-findings-from-the-gartner-customer-experience-survey/}.}
The \url{www.werk.nl} web site is run by UWV, which is the social security institute that implements employee insurances and provide labour market services to residents in the Netherlands.
Specifically, the web site supports unemployed Netherlands' residents in the process of job reintegration. Once logged in the web site, people can upload their own CVs, search for suitable jobs and, more in general, interact with UWV via messages as well as they can ask questions, file complaints, etc.
The \url{www.werk.nl} web site is structured into sections of pages and logged-in users can arbitrary switch from one to another. However, to improve the experience, it would be worthwhile introducing supporting wizards. The starting pointing for designing such wizards is to gain insights into the typical ways in which the web site is actually used.

Publicly available is an event log that collects the browsing behavior of the logged-in visitors in the period from July, 2015 to February, 2016.\footnote{The dataset is available at \url{https://doi.org/10.4121/uuid:01345ac4-7d1d-426e-92b8-24933a079412}} The event log is composed by 335655 events divided in 2624 traces.
We tried to discover a model of the web-site interaction without abstracting the event log.
Figure~\ref{fig:UWVNoAbstractionModel} shows the result obtained through the new Heuristic Miner~\cite{Mannhardt.2017e}. Similar results are also obtained through other miners and all show the problems mentioned above: the model is overcomplex, with an ``ocean'' of activity dependencies.
While this is certainly not surprising because of the freedom of visiting the web site, still one wants to discover a model that provides insights for the stakeholders.}

\begin{figure}[t]
  \centering
  \includegraphics[width=1.0\textwidth]{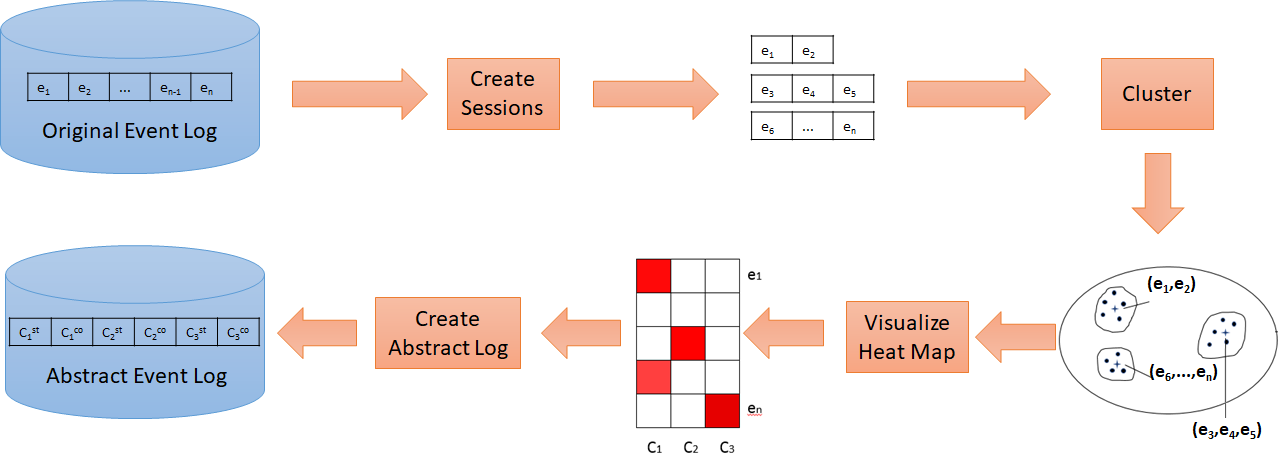}
  \caption{The steps of the abstraction technique based on sessions.}
  \label{fig:AbstractionMethodology}
\end{figure}

\section{Session-based Event-log Abstraction Technique}
\label{sec:clustering}

This section introduces the technique of clustering low-level events into high-level activities.
This procedure consists of four main steps, as visualized in Figure \ref{fig:AbstractionMethodology}. The starting point is an event log. All the traces of the event log are split into sessions, which are then clustered; the centroids of the found clusters are visualized on a heat map to provide support to assign a name to each cluster. Finally, the abstract event log is created: each session is replaced by two events (e.g.\ $C_1^{st}$ and $C_1^{co}$ in figure) of the same name as that given to the cluster to which the session belongs. The two events refer to the starting and the completion of the session and, respectively, take on the timestamps of the first and the last event of the session.

\subsection{Preliminaries}
\label{sec:prelim}

The starting point of our technique is an event log, which consists of a set of traces, each of which is a sequence of unique events:
\begin{definition}[Event, Trace, Log]
\label{def:eventLog}
Let $\ME$ be the universe of events. A \emph{\textbf{trace}} $\sigma \in \ME^{*}$ is a sequence of events. An \emph{\textbf{event log}} $\ML$ consists of a set of traces, i.e. $\ML\in \ME^{*}$.
\end{definition}
Events carry on information: given an event $e \in \ME$, $\lambda_A(e)$ and $\lambda_T(e)$ respectively return the activity associated with $e$ and the timestamp when event $e$ occurred. In the remainder, $e \in \ML$ indicates that there is a trace $\sigma \in \ML$ s.t. $e\in\sigma$.
Given a trace $\sigma'=\langle e_1,\ldots,e_n \rangle$, $\sigma'(i)$ returns the $i$-th event of the trace, namely $\sigma'(i)=e_i$; also, $|\sigma'|$ returns the number of  $\sigma'$, namely $n$.

Furthermore, given a second trace $\sigma''=\langle f_1,\ldots,f_m \rangle$, $\sigma' \oplus \sigma''$ indicates the trace obtained by concatenating $\sigma''$ at the end of of $\sigma'$, i.e.\
$\sigma' \oplus\sigma''=\langle e_1,\ldots,e_n,f_1\ldots,f_m \rangle$.

As mentioned in Section 1, we leverage on clustering techniques.
In a nutshell, these take a multiset of $n$-ples, elements of domain $D_1 \times\ldots \times D_N$, and split it into a number of disjoint smaller multisets:\footnote{Given a (multi)set $M$, $\wp(M)$ denotes the powerset, namely the set of all sub(multi)sets of $M$. The operator $\uplus$ denotes the union of multisets, namely such that the cardinality of an element in the union is the sum of the cardinality of all elements of the joined multisets.}
\begin{definition}[Clustering]
Let $\mathbb{M}$ be the set of all multisets of all data points defined over the cartesian product $D_1 \times\ldots \times D_N$.
A clustering technique can be abstracted as function $\cluster : \mathbb{M} \rightarrow \wp(\mathbb{M})$ that, given a multiset $M \in \mathbb{M}$, returns a $M$'s clustering into a set $\{M'_1,\ldots,M'_n\}$ of multisets such that $M'_1 \uplus \ldots \uplus M'_n = M$ and, for any $1 \leq i \leq j \leq n$, $m \in M'_i \land m \in M'_j \Rightarrow M'_i = M'_j$.
\end{definition}

\subsection{Creation of Sessions}

The first step of the technique is to identify the sessions. We introduce a \textbf{session threshold} $\Delta$, a time range. For each trace $\sigma = \langle e_1, \ldots, e_n \rangle$ in an event log, we iterate over its events and create a sequence of sessions $\langle s_1,\ldots,s_m \rangle$. We create a session $s_k=\langle e_i, \ldots, e_j \rangle$, subsequence of $\sigma$, if \textbf{\emph{(1)}} the timestamp's difference between $e_{i}$ and $e_{i-1}$ and $e_{j}$ and $e_{j+1}$ is larger than or equal to $\Delta$ and \textbf{\emph{(2)}} the timestamp's difference between two consecutive events in $\langle e_i, \ldots, e_j \rangle$ is smaller than $\Delta$:
\begin{definition}[Sessions of a Trace]\label{def:sessions}
Let $\sigma = \langle e_1, \ldots, e_n \rangle \in \ME^{*}$ be a log trace. Let $\Delta$ be a time interval. $\session_\Delta(\sigma)=\langle s_1,\ldots,s_m \rangle \in (\ME^*)^*$ denotes the session sequence of $\sigma$: \textbf{(1)} for any $1 \leq i < m$, $\lambda_T(s_{i+1}(1))-\lambda_T(s_{i}(|s_{i}|)) \geq \Delta$, and \linebreak \textbf{(2)}, for any $1 \leq i \leq m$ and $1 \leq j < |s_i|$, $\lambda_T(s_i(j+1)) - \lambda_T(s_i(j)) < \Delta$, and \textbf{(3)} $\sigma = s_1 \oplus \ldots \oplus s_n$.
\end{definition}
The third condition states that, if we concatenate the sessions in which $\sigma$ was split, we obtain $\sigma$ back.
The following example further clarifies:
\begin{example}
\label{ex:createSess}
Consider a trace $\sigma = \langle a_1, b_3, c_4, a_{10}, d_{13} \rangle$ of an event log $\overline{\ML}$. The letter indicates the activity name, and the subscript is the timestamp of the event's occurrence (e.g.\ d occurred at time 13). Assume that the time interval $\Delta = 5$.
One can easily see that the time difference between the second occurrance of $a$ and the first of $e$ is greater than the given time interval $\Delta$ ($\lambda_T(a_{10}) - \lambda_T(c_4) = 6 > \Delta = 5$), thus resulting in two sessions:
$\session_\Delta(\sigma) = \langle s_1, s_2 \rangle$
where $s_1 = \langle a_1, b_3, c_4 \rangle$ and $s_2 = \langle a_{10}, d_{13} \rangle$. Note that the concatenation results in $\sigma$: $\sigma = s_1 \oplus s_2$.
\end{example}

\subsection{Clustering of Sessions}
\label{sec:createSess}

Once the sessions are identified, the next step is to cluster them. To apply clustering techniques, each session needs to be encoded as a vector $p$, point of a cartesian space \mbox{$D_1 \times \ldots \times D_n$}. This encoding can be made using different policies. As an example, a session $s$ can be encoded into a vector that contains one integer dimension for each activity $a$ that, as value, takes on the number of occurrence of events referring to the activity $a$ in session $s$.
The encoding is abstracted as function $\encode(s,\sigma,\ML)$ that returns a tuple that encodes a session $s$ of a trace $\sigma$ of an event log $\ML$.
Given an event log $\ML$, we create the multiset of data points as follows:
\begin{equation}
M_\ML = \displaystyle \uplus_{\sigma \in \ML} ( \uplus_{s \in \session(\sigma)} \encode(s,\sigma,\ML) )
\label{equ:clustering}
\end{equation}
which are then clustered into $\{M'_1,\ldots,M'_n\}=\cluster(M_\ML)$.
The remainder illustrates two encodings that are of more general applicability. However,  it is possible to seamlessly plug in new encodings.

\noindent\textbf{Frequency-based Encoding.}
Let $\ML$ be an event log defined over an activity set $A_\ML=\cup_{e \in \ML} \lambda_A(e)$. Given a session $s$ of a trace $\sigma\in\ML$, the frequency-based encoding \onlyReport{returns a tuple where each of its elements is associated with a different activity of $\ML$ and takes on a value that is the number of occurrence of the respective activity in the event log.}
\onlyPaper{\linebreak $\encode_{freq}(s,\sigma,\ML)$ returns a tuple $(c_{a_1},\ldots,c_{a_n})$ where, for all $a_i \in A_\ML$, $c_{a_i}= |\{ e \in s. \lambda_A(e)=a_i\}|$ is the number of events for $a_i$ in $s$.}
For instance, the sessions $s_1$ and $s_2$ of Example~\ref{ex:createSess} at page~\pageref{ex:createSess} are encoded as quadruples where the elements from the first to the fourth dimension take on values equal to the number of occurrences of respectively $a, b, c, d$: namely, $\encode_{freq}(s_1,\sigma,\overline{\ML})=(1,1,1,0)$ and $\encode_{freq}(s_2,\sigma,\overline{\ML})=(1,0,0,1)$.
This encoding is useful when one wants to cluster on the basis of the frequency of occurrence of activities in sessions. Consider, for instance, an online retail shop where each log trace contains one event for each item of product that is added to the basket. Each web-site visit corresponds to a session. The frequency-based encoding makes a vector out of each session with as many dimensions as the products that can be potentially added to a basket: the value of a certain dimension coincides with the quantity bought of the product associated with that dimension.
\onlyReport{More formally:
\begin{definition}[Frequency-based Encoding]
Let $\ML$ be an event log; let $A = \{a_1,\ldots,a_n\}$ be the activities of $\ML$, namely $A = \cup_{\sigma \in \ML} \cup_{e \in \sigma} \lambda_A(e)$.
Given a trace $\sigma \in \ML$ and a time interval $\Delta$, let $s \in \session_\Delta(\sigma)$ be a session of $\sigma$.
The frequency-based encoding of $s$ is \[\encode_{freq}(s,\sigma,\ML)=(c_{a_1},\ldots,c_{a_n})\] such that, for all $1 \leq i \leq n$, $c_{a_i}$ is the number of events $e \in s$ for activity $a_i$, \mbox{$c_{a_i} = |\{ e \in s. \lambda_A(e)=a_i\}|$}.
\end{definition}}

\noindent\textbf{Duration-Based Encoding.}
Given a session $s=\langle es_1,\ldots,es_m \rangle$ of a trace $\sigma$ of event $\ML$, the duration-based encoding $\encode_{dur}(s,\sigma,\ML)$ returns a tuple $(d_{a_1},\ldots,d_{a_n})$ where, for all $a_i$ in log activity set $A_\ML$, $d_{a_i}$ returns the average duration of executions of activity $a_i$ in $s$. The average duration of $a_i$ can be computed as the average
\begin{equation}
\label{eq:durationEncoding}
\lambda_T(es_{j+1})-\lambda_T(es_{j})
\end{equation}
for all $es_{j}$ s.t.\ $j<m$ and $\lambda_A(es_{j})=d_{a_i}$. For the last event $es_m$, we compute the average duration of all executions of $\lambda_A(es_m)$ that were associated to all events in $\ML$ that were not the last in the respective sessions, i.e. for the events in $\ML$ for which Equation~\ref{eq:durationEncoding} can be computed.
For further clarification, let us again consider the sessions $s_1$ and $s_2$ of Example~\ref{ex:createSess}: they are encoded as quadruples where the elements from the first to the fourth dimension take on values equal to the average duration of activities $a, b, c, d$.
Let $\avg(c,\overline{\ML})$ and $\avg(d,\overline{\ML})$ be the average duration of $c$ and $d$ in the event log $\overline{\ML}$ of Example~\ref{ex:createSess}. For this example, we have $\encode_{dur}(s_1,\sigma,\overline{\ML})=(2,1,\avg(c,\overline{\ML}),0)$ and $\encode_{dur}(s_2,\sigma,\overline{\ML})=(3,0,0,\avg(d,\overline{\ML}))$.
Note that this way to compute is based on the idea that events record the starting of executing an activity, and none records the completion. The specific choice was driven by the analysis of the \url{www.werk.nl} web site: events are associated to starting visiting a web-site page and users remain on that page until they start visiting the next. However, new encoding can be put forward, which consider events as the execution's completion, cross information about resource utilizations and activity executions~\cite{Nakatumba}, or which are based on the exactly duration, if derivable/present in the event.

\begin{figure}[t!]
    \centering
    \subfloat[][]{
        \includegraphics[width=0.7\textwidth]{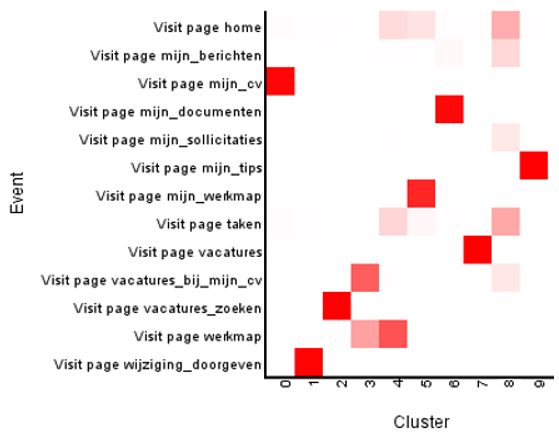}\label{fig:exampleHeatMap}
    }
    \subfloat[][]{
    \begin{minipage}[c]{0.3\textwidth}
\begin{scriptsize}
    \vspace{-7cm}
\begin{tabular}{|c|l|}
    \hline\textbf{Cluster} & \textbf{Name} \\ \hline\hline
    \multirow{ 2}{*}{0} & Visit page \\
       &  mijn\_cv \\\hline
    \multirow{ 3}{*}{1}  & Visit page \\
    &  wijzigin\_\\
    & doorgeven  \\\hline
    \multirow{ 2}{*}{2}  & Visit page \\
    &  vacatures\_zoeken  \\\hline
    \multirow{ 3}{*}{3}  & Visit page \\
    &  vacatures\_bij\_\\
    & mijn\_cv  \\\hline
    \multirow{ 2}{*}{4}  & Visit page \\
    & werkmap  \\\hline
    \multirow{ 2}{*}{5}  & Visit page \\
    &  mijn\_werkmap   \\\hline
    \multirow{ 2}{*}{6}  & Visit \\
    & mijn\_documenten  \\\hline
    \multirow{ 2}{*}{7}  & Visit page \\
    &  vacatures  \\\hline
    \multirow{ 2}{*}{8}  & Visit page \\
    & taken+home \\\hline
    \multirow{ 2}{*}{9}  & Visit page\\
    & mijn\_tips  \\ \hline
\end{tabular}
\end{scriptsize}
\end{minipage}
\label{tab:exampleNames}}
\caption{An example of heat map of the cluster centroids (part a) and of names that can be given to the clusters (part b)}
\label{fig:heatmapNames}
\end{figure}


\subsection{Visualization of Heat Maps and Creation of Abstract Event Logs}
\label{sec:heatmap}

Section~\ref{sec:createSess} produces a set of clusters $\{M'_1,\ldots,M'_n\}$ (cf.\ Equation~\ref{equ:clustering} at page \pageref{equ:clustering}), which is the input to build the abstract event log. As mentioned, clusters need to be given names. Here, we advocate the use of heatmaps to visualize the cluster centroids and, hence, facilitate the assignment of names to clusters.
An example is in Figure~\ref{fig:heatmapNames}(a), which refers to the application to the \url{werk.nl} web-site. Each row and column respectively refer to a different low-level event, dimension of the clustering space, and to a different cluster.

In particular, the centroid of each cluster is normalized between 0 and 1, and shown on the heat maps through different red-color intensities, with 0 being white and 1 being the most intense red. The color for a column X and row Y is proportional to the value of the dimension for low-level event Y in the centroid of cluster X.
The normalization of a given centroid $(c_1,\ldots,c_n)$ is achieved by dividing by the sum of the centroid's values: $(\frac{c_1}{sum},\ldots,\frac{c_n}{sum})$ where $sum = c_1 + \ldots + c_n$.
The following example well explains:
\begin{example}
\label{ex:normalization}
Let us assume the following centroids: $(1, 0, 1, 1, 0, 1)$,     $(40, 0, 2, 0, 0, 0)$,                  $(0, 0, 0, 10, 0, 1)$, $(1, 2, 0, 0, 0, 0)$, $(0, 0, 2, 2, 2, 1)$. The normalization produces \linebreak $(\frac{1}{4}, 0, \frac{1}{4}, \frac{1}{4}, 0, \frac{1}{4})$,     $(\frac{40}{42}, 0, \frac{2}{42}, 0, 0, 0)$, $(0, 0, 0, \frac{10}{11}, 0, \frac{1}{11})$,
$(\frac{1}{3}, \frac{2}{3}, 0, 0, 0, 0)$, \linebreak $(0, 0, \frac{2}{7}, \frac{2}{7}, \frac{2}{7}, \frac{1}{7})$.
\end{example}
Note that we do not normalize by simply dividing by the largest value, such as 42 in Example \ref{ex:normalization}. If we did so in Example \ref{ex:normalization},
the first, fourth and fifth centroids would be normalized to a vector with almost zero values for all dimensions.

If one obtains such a heatmap as that in Figure~\ref{fig:heatmapNames}(a), the stakeholder is largely facilitated to assign names to clusters because almost each cluster is characterized by a centroid with predominant values for one or two dimensions, each associated to a different activity. This stakeholder's involvement is optional: if this domain knowledge is absent, cluster can be given a name that just coincides with the predominant dimension or with the concatenations of those predominant.
In sum, each cluster $M_i$ is given a name $\name(M_i)$, which makes it possible to synthesize the abstract event log as follows.
\onlyReport{\begin{algorithm}[t!]
\onlyPaper{\begin{scriptsize}}
\onlyReport{\begin{footnotesize}}
\caption{Creation of an Abstract Event Log}
\label{alg:AbstractEventLog}

\KwIn{Event Log $\ML \in \ME^*$, a set $M=\{M_1,\ldots,M_n\}$ of clusters with names $\name(M_1),\ldots,\name(M_n)$}
\KwResult{Abstract Event Log}
\BlankLine
$\ML' \leftarrow \emptyset$\\
\ForEach{$\sigma \in \ML$} {
	$\sigma' \leftarrow \langle \rangle$\\
    \ForEach{\emph{\textbf{session}}$\langle e_1,\ldots,e_m\rangle \in \session(\sigma)$} {
    	$c \leftarrow \encode(\langle e_1,\ldots,e_m \rangle)$\\
        \textbf{Pick} $M_i \in M$ \textbf{s.t.} $c \in M_i$\\
        \textbf{Create Events} $e_s^{start}$ and $e_s^{complete}$ \textbf{s.t.}\\
        $\quad \lambda_A(e_s^{start})=\lambda_A(e_s^{complete})=\name(M_i)$ \\
        $\quad \lambda_T(e_s^{start})=\lambda_T(e_1)$\\
        $\quad \lambda_T(e_s^{complete})=\lambda_T(e_m)$  \\
        $\sigma' \leftarrow \sigma' \oplus \langle e_s^{start}, e_s^{complete} \rangle$ \\
    }
    $\ML' \leftarrow \ML' \cup \{ \sigma' \}$
}
\Return$(\ML')$\\
\onlyReport{\end{footnotesize}}
\onlyPaper{\end{scriptsize}}
\end{algorithm}}
\onlyReport{Algorithm~\ref{alg:AbstractEventLog} illustrates the procedure.} For each log trace $\sigma$, the algorithm builds a new trace $\sigma'$ to be added to the abstract log $\ML'$ as follows: for each session $s=\langle e_1,\ldots,e_m \rangle \in \session(\sigma)$, the algorithm determines the cluster $M_i$ to which session belongs \onlyReport{(lines 5 and 6)} and adds two events $e_s^{start}$ and $e_s^{complete}$ to the tail of $\sigma'$ \onlyReport{(lines 7 and 11)}. Events $e_s^{start}$ and $e_s^{complete}$
respectively represent the start and the end of session $s$ with the corresponding timestamps\onlyReport{(see lines 9 and 10)}, and
they refer to the high-level activity $\name(M_i)$\onlyPaper{: $\lambda_A(e_s^{start})=\lambda_A(e_s^{complete})=\name(M_i)$,
$\lambda_T(e_s^{start})=\lambda_T(e_1)$ and $\quad \lambda_T(e_s^{complete})=\lambda_T(e_m)$. The technical report~\cite{Report} that extends this paper shows the pseudo-code of the entire procedure of the log-abstraction technique.}
 \onlyReport{(line 8).}

\section{Evaluation}
\label{sec:eval}


The abstraction technique introduced in this paper has been implemented as a plug-in named \emph{Session-based Log Abstraction} in the \emph{TimeBasedAbstraction} package of the nightly-build version of ProM.\footnote{\url{http://www.promtools.org/}}
To this date, the implementation features the clustering algorithms K-means and DBSCAN algorithms.
\onlyReport{As discussed in Section~\ref{sec:createSess}, the cluster centroids are visualized on a heatmap to provide users with the necessary help to determine the high-level activity names: the heat-map visualization is provided via the \texttt{JHeatChart} library.\footnote{\url{http://www.javaheatmap.com/}}
The rest of this section illustrates the application to two case studies, for process discovery and behavior comparison.}

\subsection{Experiments on the \url{werk.nl} website}
\label{sec:werknl-experiments}

This section focuses on illustrating the successful application to the case study of the \url{werk.nl} website.
\onlyPaper{The web site supports unemployed Netherlands' residents in the process of job reintegration. Once logged in the web site, people can upload their own CVs, search for suitable jobs and, more in general, interact with UWV via messages. Also, they can ask questions, or file complaints. However, logged-in users can arbitrary switch from one page to another on the web site. The experience might likely be improved, if UWV introduced wizards for the most common ways of using the web site. This requires UWV to learn how customers visit the web site.

An event log is publicly available that collects the browsing behavior of the logged-in visitors in the period from July, 2015 to February, 2016.\footnote{The dataset is available at \url{https://doi.org/10.4121/uuid:01345ac4-7d1d-426e-92b8-24933a079412}} The event log is composed by 335655 events divided in 2624 traces. Figure~\ref{fig:UWVNoAbstractionModel} introduced the model that was discovered without log abstraction.}



To abstract the event log, we used a duration-based encoding (cf.\ Section \ref{sec:createSess}): it is certainly more important to consider how long visitors stay on a web page, rather than just counting the number of visits of the different pages. For instance, three 1-minute visits of any page should not have more weight than a 30-minute visit of the page. The session threshold $\Delta$ was set to 15 minutes, because it coincides with the timeout of \url{www.werk.nl}.

Initially, the data points that encode the sessions of the log traces were clustered via DBSCAN.
The generation of the clusters with DBSCAN took nearly 2 hours on a low-profile laptop with 8 Gb of RAM.
The clusters' centroids were visualized through the heatmap in Figure~\ref{fig:heatmapNames}(a). To help stakeholders, the plug-in removes the rows referring to low-level events that, when normalized, are associated with nearly-zero values of all centroids. The results in Figure~\ref{fig:heatmapNames}(a) are certainly very interesting: the sessions of a certain cluster are characterized by few particular pages, long and often visited. Note that DBSCAN does not always return clusters: DBSCAN would have failed, if it had not been possible to cluster the data points.
Without using additional domain knowledge, each cluster was named after the low-level event (i.e.\ web page) that refers to the dimension with the highest value in the centroid (the most intense red color). This led to the names in Table~\ref{fig:heatmapNames}(b).

Once the names are assigned to clusters, we generated an accordant, abstract event log.
To validate the quality of the abstract event log, this was randomly split into a 70\% part, which was used for discovery, and a 30\%, for testing.
The DBScan algorithm naturally computes outliers, namely points that are not assigned to any cluster.
Results show that, if those outliers are simply filtered out, the quality of the discovered model is significantly dropped (see discussion below, summarized in Table~\ref{tab:sumCCUWV}). Therefore, we performed a post-processing where each outlier session is manually inserted into the cluster with the closest centroid.
The abstract event log with the manual cluster assignment of outliers was used as input for the new Heuristic Miner~\cite{Mannhardt.2017e}, thus discovering the model in
\begin{figure}[t!]
    \centering
    \includegraphics[width=0.9\textwidth]{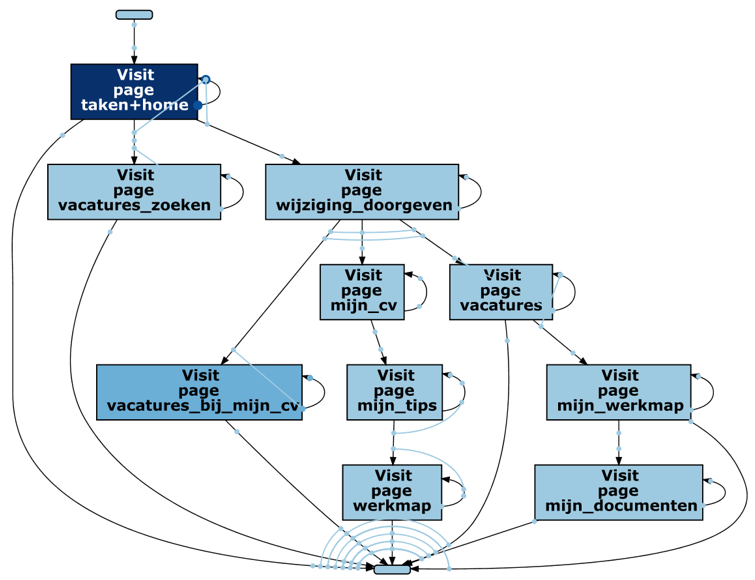}
    \caption{Process model produced by the Heuristic Miner~\cite{Mannhardt.2017e} on 70\% of the abstract event log of the \url{werk.nl} dataset, clustering via DBSCAN.}
    \label{fig:uwvDBSCAN70}
\end{figure}
Figure~\ref{fig:uwvDBSCAN70}, using the Causal-Net notation~\cite{Aalst.2016}.

\onlyReport{\begin{figure}[t!]
    \centering
    \includegraphics[width=0.87\textwidth]{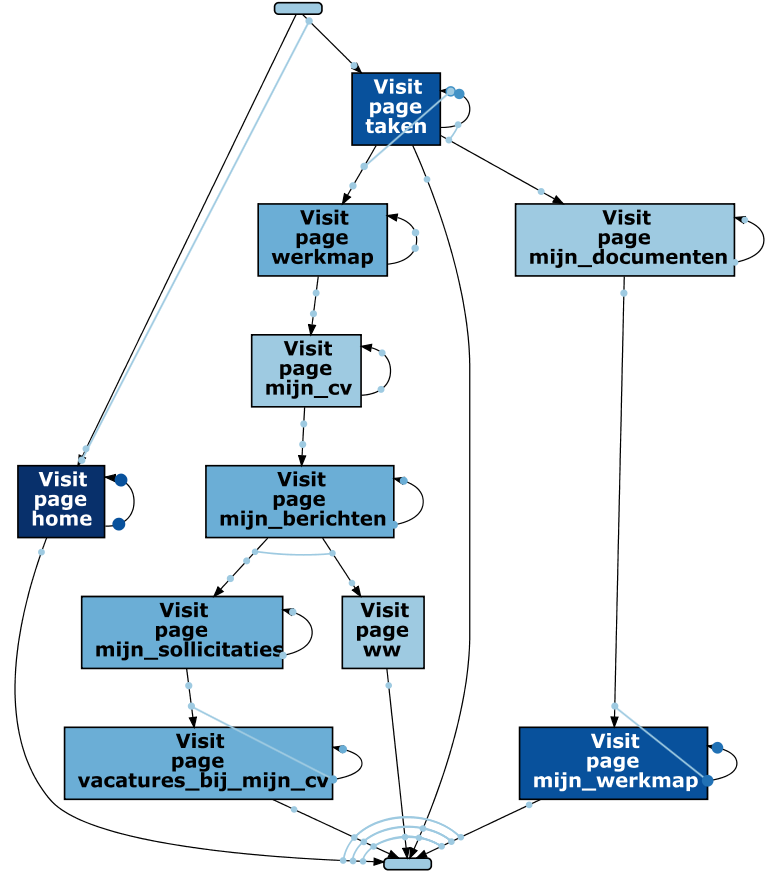}
    \caption{Process model produced by the Heuristic Miner~\cite{Mannhardt.2017e} on 70\% of the abstract event log of the \url{werk.nl} dataset, clustering via K-Means.}
    \label{fig:uwvKM70}
\end{figure}}

The same procedure was employed to discover a high-level model with K-Means, using the same 70\% of the traces for discovery, and the same temporal threshold and encoding as for DBSCAN. Note that, compared with DBSCAN, K-Means requires one to explicitly set the number of clusters to create. Our implementation features the \emph{Elbow Method} to facilitate the setup~\cite{ElbowMethod}: when applied to the case study, creating ten clusters seemed to provide a good balancing between minimizing the error and not scattering the sessions among too many clusters (i.e. high-level activities). \onlyReport{The resulting model is in Figure~\ref{fig:uwvKM70}.} \onlyPaper{The resulting model is shown in the technical report~\cite{Report}.}

The quality of these models was assessed through the classical process-mining metrics of fitness, precision, generalization and simplicity~\cite{Aalst.2016}. Fitness was computed on the 30\% of abstract log that was not used for discovery. This is accordant with typical machine-learning methods of verifying process-model ``recall'' on traces that were not used for discovery.
Conversely, precision and generalization were computed on the entire abstract log.
Finally, simplicity was measured as the sum of activities, arcs and bindings of the causal nets.
Since fitness, precision and generalization are traditionally defined on Petri nets~\cite{Aalst.2016}, causal nets were converted to Petri nets using the implementation in~\cite{Mannhardt.2017e}. The resulting Petri nets were manually adjusted to ensure soundness while not adding extra behavior. Of course, \emph{to keep the comparison fair, all models were discovered by the Heuristic Miner~\cite{Mannhardt.2017e}, using the same configuration of parameters.} This includes the model in Figure~\ref{fig:UWVNoAbstractionModel}.

\begin{table}[t!]
\centering
\onlyPaper{\begin{scriptsize}}
\begin{tabular}{|l|l|l|l|}
\hline
               & \textbf{K-Means} & \textbf{DBSCAN With Post-Processing} & \textbf{DBSCAN No Post-Processing} \\ \hline\hline
Fitness        & 0.6637     & 0.6270    & 0.2785    \\ \hline
Precision      & 0.33192    & 0.74779   & 0.68247   \\ \hline
Generalization & 0.99962    & 0.99996   & 0.99998   \\ \hline
Simplicity     & 81         & 91        & 79        \\ \hline
\end{tabular}
\onlyPaper{\end{scriptsize}}
\caption{Measures of the quality of the models discovered on the log abstracted through K-Means and DBSCAN. For the DBSCAN, we report on the values when the postprocessing to manually insert outlier was and was not performed.}
\label{tab:sumCCUWV}
\end{table}

Table~\ref{tab:sumCCUWV} illustrates the results of the comparison of the models discovered through the abstract event logs obtained via K-Means and DBSCAN. They equally generalize and are of similar complexity (variation of simplicity is around 10-12\%). The abstract model when applying DBSCAN without post processing shows very poor fitness, which is conversely satisfactory when applying K-Means or DBSCAN with post processing. Focusing on precision, the model of DBSCAN with post-processing is characterized by a precision that is 2.25 times than the precision of the K-Means model.
This leads to the conclusion that DBScan with post-processing has produced a better model, in terms of fitness, simplicity, precision and generalization. Intuitively, this is not surprising: DBScan is based on maximizing the cluster density, ensuring that ``similar'' sessions are put in the same cluster.

In conclusion, the model in Figure~\ref{fig:uwvDBSCAN70} is the most preferable, and unarguably more understandable, if compared with the non-abstract model in Figure~\ref{fig:UWVNoAbstractionModel}.
\onlyReport{From a business viewpoint, it illustrates that typical users navigate the \url{werk.nl} web site as follows.
During the first session, users visit the home page and, also, page \emph{taken} (Dutch for \emph{tasks}), where they can see the tasks assigned by UWV (e.g.\ to upload certain documents). If no tasks are assigned to do via the web site, the interaction with the web site completes. If any tasks are, users look for jobs to apply for (page \emph{vacatures\_zoeken}) and/or amend the information that they previously provided (page \emph{wijziging\_doorgeven}). If information is amended, usually an updated curriculum is uploaded (cf.\ the branch of the model starting with page \emph{mijn\_cv}) and/or the visitor looks and possibly applies for jobs (cf.\ the branches of the model starting with pages \emph{vacature} and \emph{vacature\_bij\_mijn\_cv}, which are either both executed or both skipped). Looking at statistics, the mean and median duration of the web-site interaction (i.e.\ the log traces) is around 20 weeks (more than 4 months) and, hence, the visiting sessions are certainly temporarily spread. One can also observe that}
\onlyPaper{Space limitations prevent us from a thorough discussion of the insights gained through the model about the typical navigation behavior. We refer to report~\cite{Report} for a detailed discussion. Here, we only discuss the most interesting result: }
Every session type is usually repeated multiple times, and this is likely due to the fact that the corresponding tasks are carried on through similar sessions in consecutive days. It is, however, remarkable that the model does not contain larger loops involving different session types. This means that the web site is visited in conceptual sections: when users start access pages of a given section, the pages of previous sections will no longer be visited. Note that the web site does not define sections, nor does it restrict the order with which pages can be visited. In fact, this testifies the benefits of introducing wizards. We acknowledge that information is lost in the abstraction. However, this loss is justified by gaining comprehensible business knowledge. As a matter of fact, this model was shown to one UWV's stakeholder, who literally said \emph{``this is the most understandable analysis of the web-site behavior that I have seen, certainly beyond the results seen for the BPI Challenge''}.\footnote{Indeed, the BPI challenge in 2016 was based on the same event data - \url{https://www.win.tue.nl/bpi/doku.php?id=2016:challenge}.}

\subsection{Evaluation on a Building-Permit Process}

\begin{figure}[t!]
    \centering
    \includegraphics[width=1\textwidth]{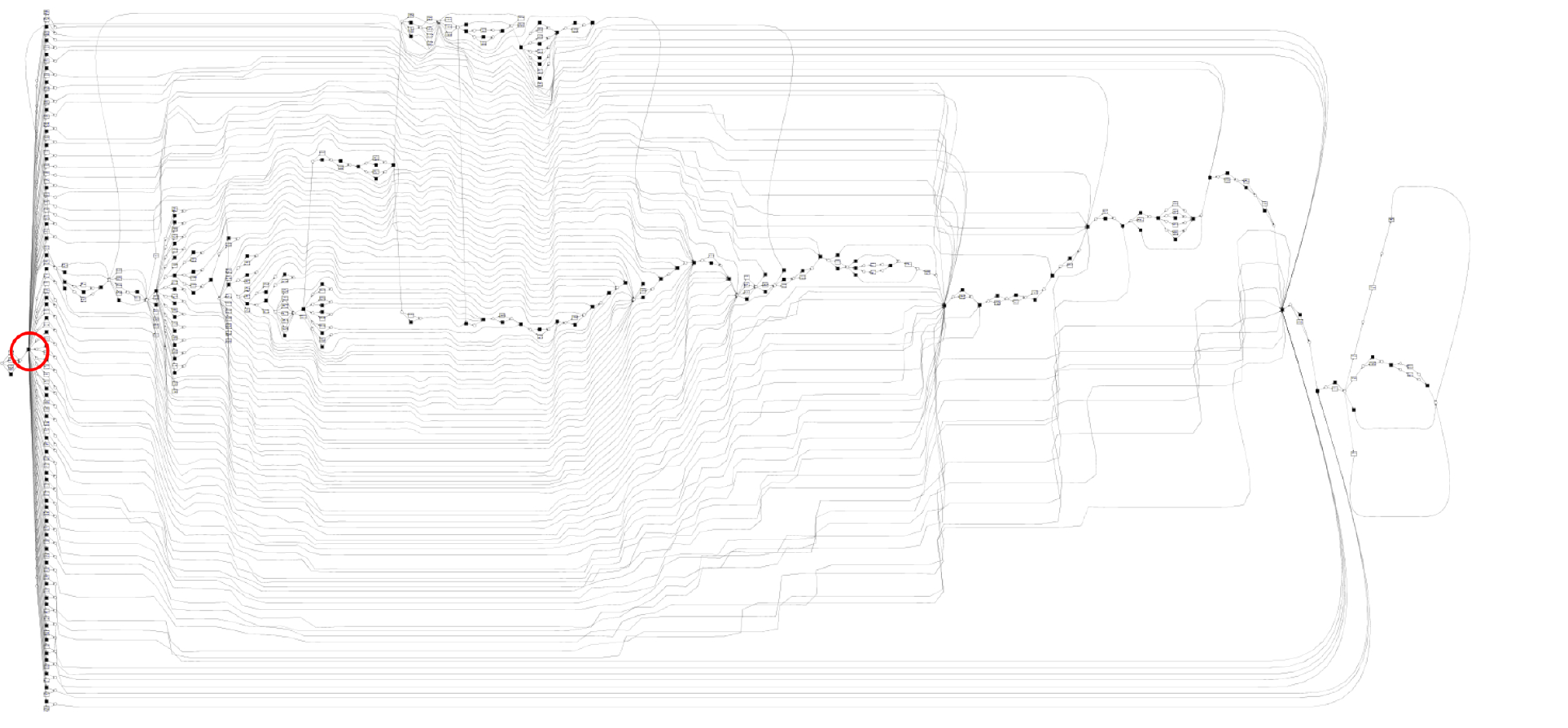}
    \caption{Building-permit process model produced by the Inductive Miner without abstraction: overly complex to be insightful.}
    \label{fig:buildingPermitNoAbstraction}
\end{figure}

This section discusses a second case study to illustrate the applicability of the technique beyond \url{werk.nl}. This case study refers to the execution of process to manage building-permit applications in a Dutch municipality.\footnote{The event log is available at {http://dx.doi.org/10.4121/uuid:63a8435a-077d-4ece-97cd-2c76d394d99c}} There are 304 different activities denoted by their respective English name as recorded in attribute \emph{taskNameEN}.
The event log spans over a period of approximately four years and consists of 44354 events divided in 832 cases.
Figure~\ref{fig:buildingPermitNoAbstraction} shows the model discovered with the \emph{Inductive Miner - Infrequent Behavior}~\cite{Leemans.2013}, using the default configuration.
The model exactly shows the same problems as that in Figure~\ref{fig:UWVNoAbstractionModel}: The large variability has made the miner discover an overly complex model. See, e.g., the large OR split around the area highlighted by a red circle in the picture.
We applied the abstraction technique to the event log, using the frequency-based encoding (cf.\ Section~\ref{sec:createSess}) and the DB-SCAN clustering algorithm with post processing, which proved to perform better for the first case study reported in Section~\ref{sec:werknl-experiments}. A session threshold of 8 hours was employed so that the events of the same day were put in the same (work) session.

The clustering step resulted in \onlyReport{the heatmap in Figure~\ref{fig:heatmapNames}(a) where, similarly to the previous case study, infrequent activities are filtered out, and} \onlyPaper{a heatmap structurally similar to Figure~\ref{fig:heatmapNames}(a) where} each cluster centroid has significantly non-zero values for the dimensions for one or few low-level activities. Analogously to the first case study, clusters were given the same name as the low-level activity with the most intense red colour in the heat map, possibly concatenated with the names of the additional activities with a significantly red color\onlyPaper{.} \onlyReport{(see
Table~\ref{tab:buildingpermitNames}).} \onlyPaper{The report~\cite{Report} shows the heatmap and the table that maps clusters to name.}
\onlyReport{\begin{figure}[t!]
    \centering
        \subfloat[][]{
        \includegraphics[width=0.8\textwidth]{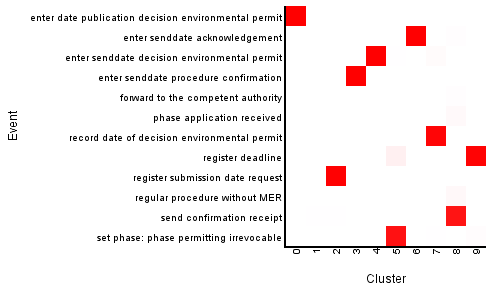}\label{fig:permitHeatMap}
    }

    \subfloat[][]{
    \begin{minipage}[c]{0.7\textwidth}
\begin{scriptsize}
\begin{tabular}{|c|l|}
    \hline\textbf{Cluster} & \textbf{Name} \\ \hline\hline
    0  & enter date   publication decision environmental permit   \\\hline
    1  & completed subcases  content  \\\hline
    2  & register submission date request  \\ \hline
    3 & enter senddate procedure confirmation \\\hline
    4 &enter senddate   decision environmental permit  \\\hline
    5  & set phase: phase  permitting irrevocable   \& register deadline  \\\hline
   6  & enter    senddate  acknowledgement  \\\hline
    7  & record date of decision  environmental permit   \\\hline
    \multirow{ 2}{*}{8}  & forward to the compotent authority \& send confirmation receipt \\
    &  \& regular procedure without MER   \& phase application received  \\\hline
    9  & register deadline \\\hline
\end{tabular}
\end{scriptsize}
\end{minipage}
\label{tab:buildingpermitNames}}
\caption{The heat map of the cluster centroids for the building-permit process (part a) and the names given to the clusters (part b)}
\end{figure}}

\begin{figure}[t!]
    \centering
    \includegraphics[width=1\textwidth]{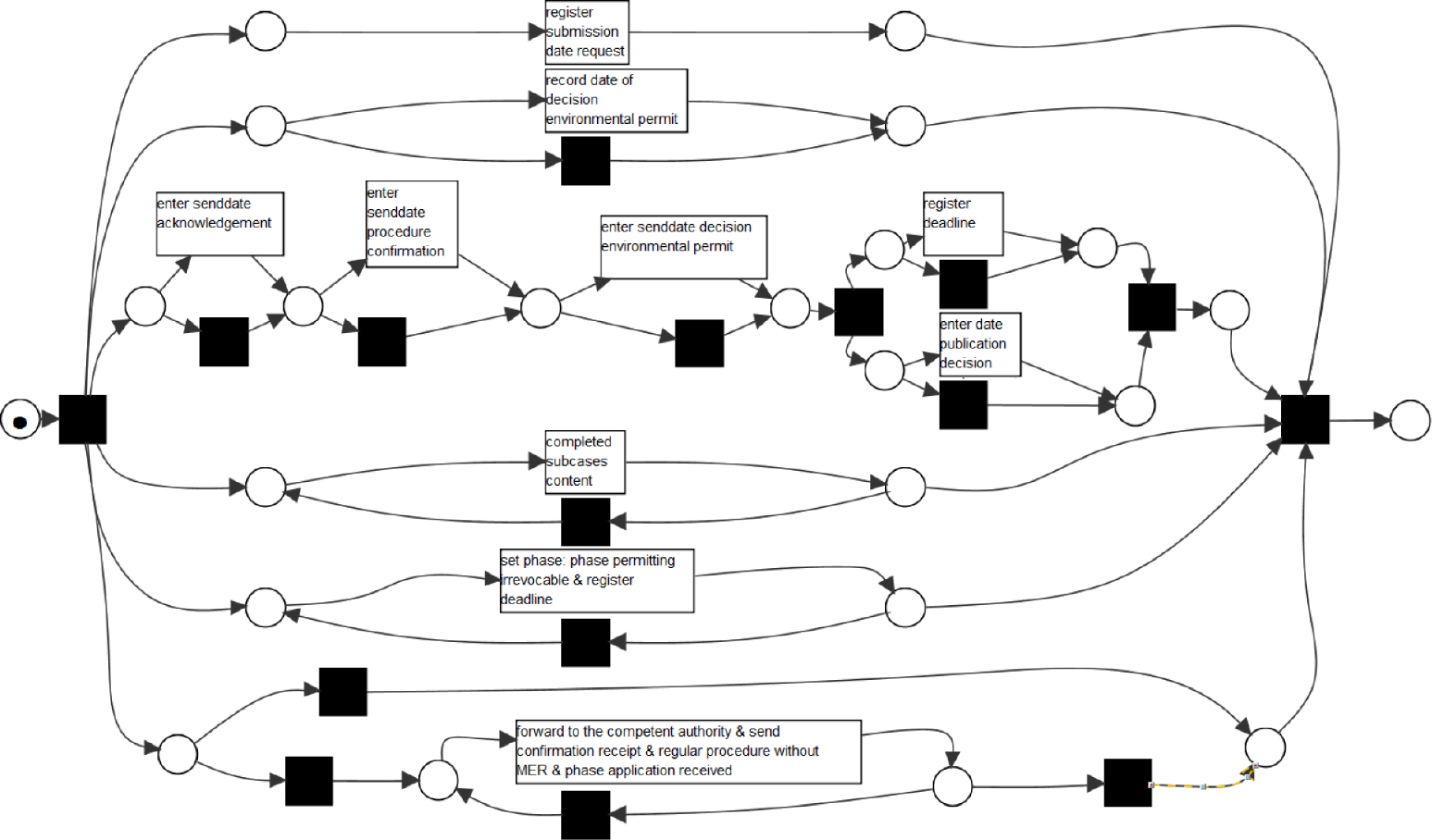}
    \caption{Building-permit process model produced by the Inductive Miner with abstraction, clustering via DBSCAN. The fitness value is 0.79; the precision and generalization values are not reported because the reference software implementation never terminated.}
    \label{fig:buildingPermitAbstraction}
\end{figure}

The abstract event log was then generated and used as input for the \emph{Inductive Miner - Infrequent} with default parameter values, namely the same as for the not-abstracted model in Figure~\ref{fig:buildingPermitNoAbstraction}. This yielded the model in Figure~\ref{fig:buildingPermitAbstraction}, which is unarguably simplified, emphasising the most salient behavioral aspects. This model is a good representation of the actual behavior: its fitness is 0.79. Unfortunately, it was not possible to compute precision and generalization because the reference ProM implementation (see~\cite{Aalst.2016}) got stuck and never terminated the computation.

\begin{figure}[t!]
    \centering
    \includegraphics[width=1\textwidth]{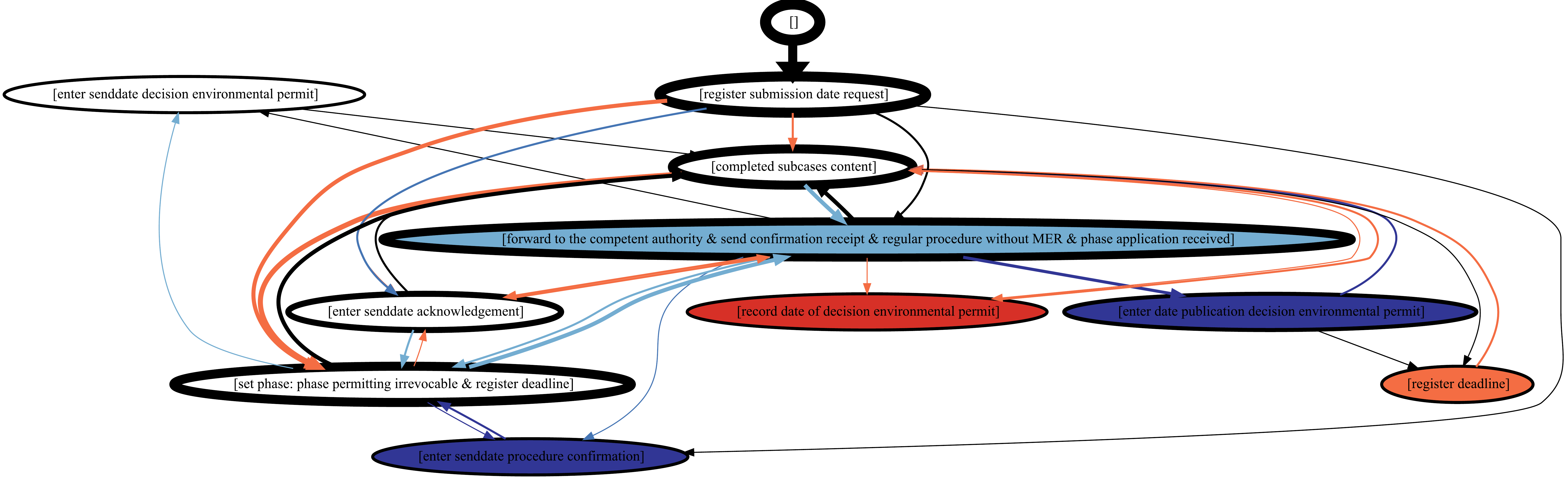}
    \caption{Comparison of the building-permit process behavior between executions when resource 560458 is responsible and when others are.}
    \label{fig:comparison}
\end{figure}

We previously claimed that event abstraction is not only about model discovery, and it enables a fruitful application of a large repertoire of process-mining techniques. The remainder will provide a support to this claim. In particular, we will show that it makes it possible to highlight that the executions under the responsibility of certain resources are statistically different from those of other resources. To achieve this, we leveraged on the technique proposed in~\cite{10.1007/978-3-319-39696-5_10}. The technique allowed us (1) to find out that the executions under the responsibility of resource 560458 are remarkably different, and (2) to pinpoint what these differences are.
The latter piece of knowledge can be gained by looking at the transition system obtained by the technique in~\cite{10.1007/978-3-319-39696-5_10}, which is in Figure~\ref{fig:comparison}. In the transition system, nodes are the event's activities and an arc between two activity nodes indicated that the event log shows that sometimes
the source activity is followed by the destination activity. Nodes and arcs are coloured with different shades of blue and orange to indicate  that the activity or transition is statistically more or is less frequent for 560458, respectively. The thickness of arcs and node's borders signifies the frequency of occurrence. The colour's darkness is proportional to the average difference. In Figure~\ref{fig:comparison}, e.g., high-level activity \emph{entersend date procedure confirmation} stands out: it occurs in 67\% of the cases of resource 560458 versus 13.7\% of the cases where others are responsible. Similar proportions are also observed in high-level \emph{enter date publication decision environmental permit}. Conversely, high-level \emph{register deadline} is colored orange, showing that it is statistically more frequent for the cases in which other resources than 560458 are responsible.
It follows quite naturally that, without abstraction, the behavior complexity represented in Figure~\ref{fig:buildingPermitNoAbstraction} would generate such a complex transition system that no fruitful insights could be derived. 

\section{Related Work}
\label{sec:relatedwork}

A large body of research has been conducted on log abstraction. It can be grouped in two categories: supervised and unsupervised abstraction. The difference is that supervised abstraction techniques require process analysts to provide domain knowledge, while unsupervised does not rely on additional information.

\noindent\textbf{Supervised Abstraction Methods.} Baier et al.\ provide a number of approaches that, based on some process documentation, map events to higher-level activities\onlyPaper{~\cite{Baier2015PhD}}\onlyReport{~\cite{Baier2015PhD,Baier2013BridgingAbstractionLayers,SAC-2015-0001RMW}}, using log-replay techniques and solving constraint-satisfaction problems.
The idea of replaying logs onto partial models is also in~\cite{mannhardt2016low}: the input is a set of models of the life cycles of the high-level activities, where each life-cycle step is manually mapped to low-level events.
Ferreira et al.~\cite{ferreira2013mining} rely on the provision of one Markov model, where each Markov-model transition is a different high-level activity. \onlyReport{In turn, each transition is broken down into a new Markov model where low-level events are modelled.}
Fazzinga et al.~\cite{fazzinga2015probabilistic} assume process analysts to provide a probabilistic process model with the high-level activities, along with a probabilistic mapping between low-level events and high-level activities. It returns an enumeration of all potential interpretations of each log traces in terms of high-level activities, ranked by the respective likelihood.
In~\cite{T_S_H_vdA@INTELLISYS2016}, authors propose a supervised abstraction technique that is applicable in those case in which annotations with the high-level interpretations of the low-level events are available for a  subset  of  traces.

\noindent\textbf{Unsupervised Abstraction Methods.} Log abstraction is related with episode mining and its application to Process Mining (a.k.a.\ discovery of local process models)~\cite{DBLP:conf/simpda/LeemansA14a,TAX2016183}). In fact, Mannhardt and Tax propose a method that combines local process model discovery with the supervised abstraction technique in~\cite{mannhardt2016low}. However, the technique relies on the ability to discover a limited number of local process models that are accurate and cover most of the low-level event activities.
In~\cite{gunther2009activity}, G{\"u}nther et al.\ cluster events looking at their correlation, which is based on the vicinity of occurrences of events for the same low-level activity in the entire log. Clustering is also the basic idea of~\cite{Dongen2009Fuzzy} to cluster events through a fuzzy k-medoids algorithm.
Both~\cite{Dongen2009Fuzzy} and~\cite{gunther2009activity} share the drawback that the time aspects are not considered and, thus, they can cluster events that are temporarily distant (e.g.\ web-site visits that are weeks far from each other).
Also, \cite{Dongen2009Fuzzy} only aims to discover a fuzzy high-level model, instead of abstracting event logs to enable a broader process-mining application, whereas~\cite{gunther2009activity} assumes a transitive nature of the property of activity correlation, which does not always hold. See Figure~\ref{fig:heatmapNames}(a): cluster 3 shows a correlation between \emph{Visit page werkmap} and \emph{Visit page vacature\_bij\_mijn\_CV} and cluster 4 shows a correlation between \emph{Visit page werkmap} and \emph{Visit page taken}, while no correlation exists between \emph{Visit page vacature\_bij\_mijn\_CV} and \emph{Visit page taken}.
Finally, van Eck et al.~\cite{Eck2016} illustrate a technique to gather observations from sensor data, encode and cluster them in a similar way as our approach does. However, they assume that events (in fact, sensor observations) are generated at a constant rate.

\noindent \textbf{Log Clustering vs Log Abstraction.} This paper has discussed a log-abstraction technique that builds on machine-learning clustering techniques. Event-log clustering also leverages on the same techniques~\cite{DeWeerdt2018}. However, event-log clustering has a different purpose because it is based on the idea to split the traces into homogenous groups, without altering the contents of the traces themselves.

\section{Final Remarks}
\label{sec:conclusion}

Abstracting and grouping low-level events to high-level activities is a problem that is receiving a lot of attention.
Often, event logs are not immediately ready to be used because they model concepts that are not at the right business level and/or they exhibit a too broad variety of behavior to be summarized into one simple model, map, diagram, etc.

Section \ref{sec:relatedwork} illustrates how, on the one hand, supervised methods often require vast domain knowledge (e.g.\ through process models, Markov chains or mapping ontologies), which is not always possible to provide. On the other hand, unsupervised methods show limitations, related to the absence of any external knowledge. This paper reports on a third way where very limited domain knowledge is necessary.
The technique is based on the idea that a trace can be regarded as a sequence of sessions each of which terminates when no additional events occur within a user-defined time interval. The sessions are later clustered; finally, a heatmap visualization of the clusters is provided to domain experts, so that they could assign meaningful high-level concepts to the sessions, i.e.\ sequences of low-level events.
Admittedly, the concept of sessions and the use of clustering techniques and heat-map visualizations are not novel in process mining, if each is taken in isolation.
The innovation here is that we ensemble them to provide a solution to the problem of abstracting low-level events to high-level concepts, with the advantages mentioned above.

Section~\ref{sec:eval} reports on the successful evaluation of the proposed technique on two case studies, discussing both a quantitative and a qualitative analysis.
The qualitative analysis shows that our log-abstraction technique (1) allows overly complex models to be simplified, by focusing on the higher-level concepts, and (2) is applicable beyond process discovery (see Figure~\ref{fig:comparison}).
Quantitatively, the evaluation showed that the discovered models can reasonably balance the typical process-mining metrics: fitness, precision, generalization and simplicity~\cite{Aalst.2016}.

The technique does not depend on any clustering algorithm, and this explains why concrete algorithms are only mentioned in Section~\ref{sec:eval}.
While we acknowledge that a more thorough assessment is necessary, Section~\ref{sec:eval} shows that the best performances are with DBScan, which has the advantage of automatically computing the best number of clusters.
This motivates further that our technique requires to supply little knowledge: considering that the provision of the cluster names is optional (cf.\ Section~\ref{sec:heatmap}), the only input for our technique is the session threshold.

The technique is applicable to all those domain fields where customers perform activities in batches/sessions, including retail shopping (e.g.\ Amazon or supermarkets), health care (e.g.\ hospitals) as well as scenarios of home automation and/or IoT.
All of these domains loosely constrain the order with which the activities are executed, which ultimately leads to an ``ocean'' of alternative behavior.

In spite of the assessment reported in this paper, we acknowledge that further validation is needed, especially in such domains as those mentioned in the previous paragraph.
In parallel, the technique can be further extended towards achieving better clustering.
Firstly, the technique needs to be extended to consider the entire event payload, instead of just limiting to the sole activity names. For instance, for the \emph{werk.nl} case study, one could add clustering dimensions related to the customer age, gender, geographic locations, etc., providing extra information towards a more accurate clustering. This is also very relevant for such domains as domotics, where, e.g., the use of an oven at 180 ${}^\circ{C}$ may be conceptually different than using it at 240 ${}^\circ{C}$.
Secondly, we plan to explore hierarchical clustering because it would allow one to tune the level of aggregation that is achieved through the log abstraction. Thirdly, the number of low-level activities is generally large. Therefore, it is worth investigating the benefits, if any, of reducing the low-level activities to consider when applying clustering.

Last and not least, we aim to specialize the general technique for the discovery of hierarchical processes and analyze the structure of the sessions within each separate cluster. The sessions within each cluster can be seen as traces of a sub log, which can be used as input to discover small (fragments of) models, to be later combined with the model discovered via abstract event logs.

\bibliographystyle{splncs04}

\end{document}